\documentclass[10pt,twocolumn,letterpaper]{article}

\usepackage[pagenumbers]{cvpr} 

\usepackage[dvipsnames]{xcolor}

\definecolor{cvprblue}{rgb}{0.21,0.49,0.74}
\usepackage[pagebackref,breaklinks,colorlinks,citecolor=cvprblue]{hyperref}

\def\paperID{286} 
\def\confName{3DV\xspace}
\def\confYear{2025\xspace}

\title{OpticFusion: Multi-Modal Neural Implicit 3D Reconstruction of Microstructures by Fusing White Light Interferometry and Optical Microscopy}

\author{
{Shuo Chen \;
Yijin Li \;
Guofeng Zhang$^\dagger$}\\
{State Key Lab of CAD\&CG, Zhejiang University}\\
{
\tt\small \{{chenshuo.eric}, {eugenelee}, {zhangguofeng}\}@{zju.edu.cn}
}\\
}

\begin{document}
\maketitle
\def\thefootnote{$^\dagger$}\footnotetext{Corresponding Author}
\begin{abstract}
White Light Interferometry (WLI) is a precise optical tool for measuring the 3D topography of microstructures. However, conventional WLI cannot capture the natural color of a sample's surface, which is essential for many microscale research applications that require both 3D geometry and color information. Previous methods have attempted to overcome this limitation by modifying WLI hardware and analysis software, but these solutions are often costly. In this work, we address this challenge from a computer vision multi-modal reconstruction perspective for the first time. We introduce OpticFusion, a novel approach that uses an additional digital optical microscope (OM) to achieve 3D reconstruction with natural color textures using multi-view WLI and OM images. Our method employs a two-step data association process to obtain the poses of WLI and OM data. By leveraging the neural implicit representation, we fuse multi-modal data and apply color decomposition technology to extract the sample's natural color. Tested on our multi-modal dataset of various microscale samples, OpticFusion achieves detailed 3D reconstructions with color textures. Our method provides an effective tool for practical applications across numerous microscale research fields. The source code and our real-world dataset are available at \href{https://github.com/zju3dv/OpticFusion}{https://github.com/zju3dv/OpticFusion}.
\end{abstract}    
\section{Introduction}
\label{sec:intro}
\begin{figure*}[h] 
    \centering
    \includegraphics[width=\linewidth]{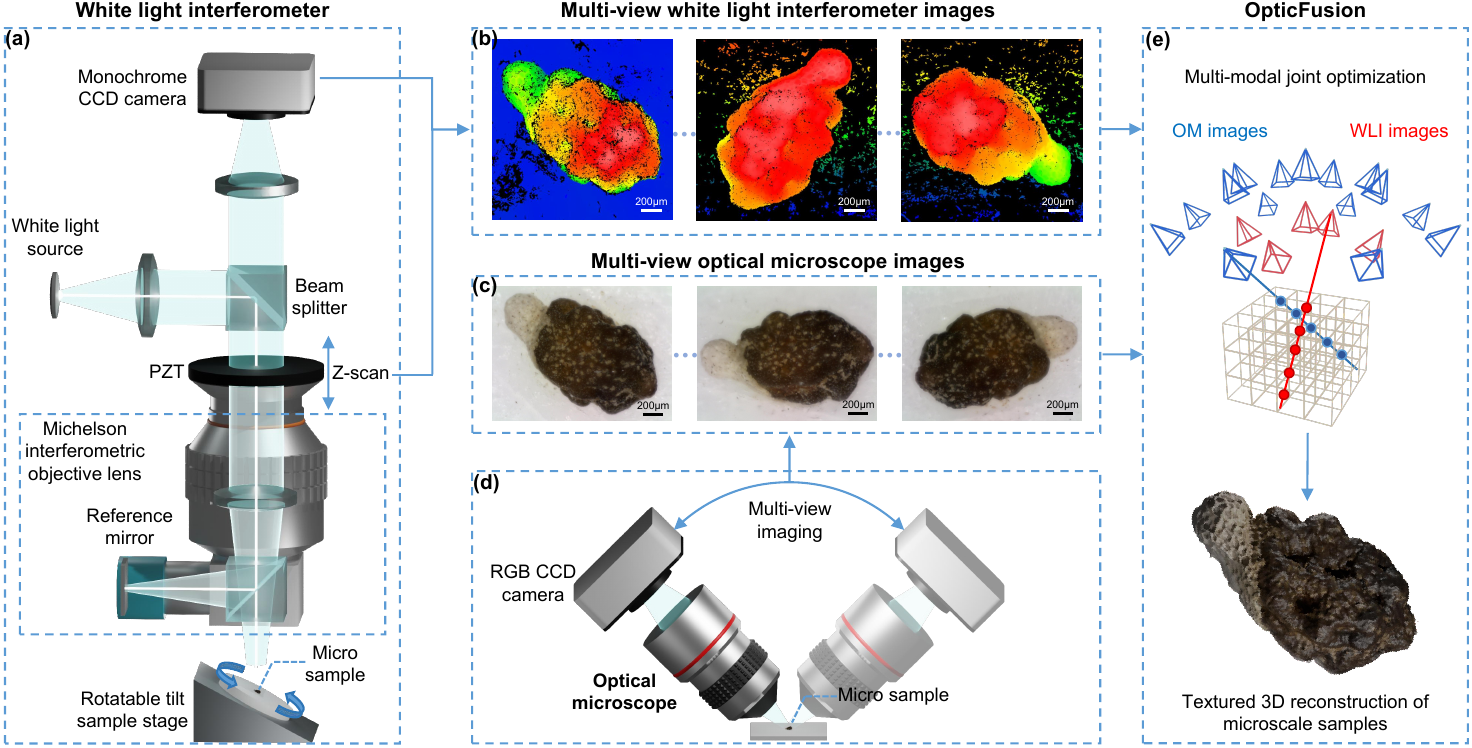}
    \caption{ \textbf{Multi-modal neural implicit reconstruction of microscale samples with multi-view images from white light interferometer and optical microscope.} \textbf{(a)} The principle of multi-view WLI scanning. \textbf{(b, c)} The output of white light interferometer and optical microscope. \textbf{(d)} The multi-view imaging with an optical microscope. \textbf{(e)} OpticFusion reconstructs 3D models of microstructures with their natural surface colors, without costly modifications of the WLI system.}
    \label{fig: data_capture}
    \vspace{-1em}
\end{figure*}

The 3D structure of microscale samples is crucial in numerous scientific research fields and industrial production processes. White light interferometry (WLI)~\cite{WLI, CSI} is widely used for high-precision 3D surface topography measurement, offering a lateral resolution of approximately 0.5 micrometers and a vertical accuracy at the sub-nanometer scale across diverse surfaces. Consequently, WLI is utilized in ultra-precision machining~\cite{li2018calibration, zhang2019manufacturing, lucca2020ultra}, integrated circuit inspection~\cite{davidson1987application, hirabayashi2002fast, roy2002geometric}, biological structure analysis~\cite{spider_WLI, biological1, biological2}, and other applications. 
In WLI (\cref{fig: data_capture}\textcolor{red}{a}), broadband white light with a very short coherence length serves as the light source. A beamsplitter divides the light into a reference beam and a measurement beam, directed toward a reference mirror and the sample surface. High-contrast interference fringes are captured by a charged-coupled device (CCD) camera only when the optical path difference (OPD) is close to zero. As the piezoelectric tube (PZT) scans along the Z-axis, the zero-OPD position shifts according to the sample's 3D structure. By determining the Z-value of the PZT that maximized the interference intensity at each point on the sample surface during the scanning, the corresponding 3D topography of the microstructure's surface can be obtained (\cref{fig: data_capture}\textcolor{red}b).

Although WLI uses white light to illuminate the sample, conventional WLI instruments typically employ monochrome CCD cameras to capture grayscale interference patterns, resulting in the loss of the sample's color information. However, the natural color of the sample can be valuable in many applications, such as observing color changes on laser-treated metal surfaces~\cite{antonczak2013laser}, identifying defects and corrosion on samples~\cite{song2010surface}, and distinguishing boundaries between different materials~\cite{huang2023vertical}.
Currently, several methods exist to obtain color images in WLI. One direct approach is to use a three-chip color CCD camera~\cite{three_chip_CCD} or a single-chip color CCD camera with a Bayer filter~\cite{one_chip_CCD1, one_chip_CCD2}. However, this approach significantly impacts WLI's lateral resolution, accuracy, and throughput~\cite{multi_light}. Another method employs switchable RGB light sources~\cite{multi_light}. Generally, these approaches require modifications to the WLI hardware and interference pattern analysis software, making them impractical for existing commercial monochromatic WLI instruments. An alternative approach~\cite{fourier1, fourier2} utilizes a reference sample with known reflectance and Fourier spectral analysis of the monochromatic interference patterns to obtain the sample's reflectance spectrum. Nonetheless, some commercial WLI instruments only provide the final topography map and do not allow users to access the raw interference patterns. 
In this paper, we propose a novel approach that does not rely on costly hardware modifications to WLI. For the first time, we introduce a multi-modal 3D reconstruction method utilizing only a commercial WLI instrument and a digital optical microscope (OM, \cref{fig: data_capture}\textcolor{red}d). This method enables accurate 3D reconstruction of microscale samples with their natural surface colors.

We propose OpticFusion, a method for neural implicit surface reconstruction of microscale samples using multi-view WLI (\cref{fig: data_capture}\textcolor{red}b) and OM images (\cref{fig: data_capture}\textcolor{red}c). 
Unlike typical multi-sensor systems with data alignment or fixed relative poses, our approach acquires WLI and OM images independently.
Therefore, we employ a two-step data association method to determine the poses of all WLI and OM images within an absolute-scale coordinate system.
In the reconstruction process, we jointly optimize a neural network-based Signed Distance Function (SDF) using both multi-view WLI and OM images  (\cref{fig: data_capture}\textcolor{red}e). This approach allows OM images to fill voids where WLI data lack measurements, enhancing the reconstruction quality and producing complete, detailed surface models of microscale samples.
Notably, in microscale research, the focus is not on synthesizing novel views but on capturing the intrinsic colors of the sample itself, independent of the view direction. 
To achieve this, we incorporate color decomposition strategies from Intrinsic-NeRF~\cite{intrinsicnerf} and Color-NeuS~\cite{color_neus}, adding a view-dependent residual branch into the color network, which results in more natural colors for the model texture.

To validate the effectiveness of our method, we collect a real dataset consisting of multi-view WLI and OM images of samples, with surface feature sizes ranging from tens to hundreds of micrometers, including a butterfly wing, flower seeds, a circuit board, and a microsensor. The results show that OpticFusion accurately reconstructs the surface details of various samples while texturing natural colors to the model. Additionally, we simulate the imaging characteristics of WLI to create a synthetic dataset. 
We demonstrate the superiority of combining WLI and OM data by quantitatively evaluating the reconstruction quality.
We also show that our reconstructed results can be used for roughness measurement of microscale surfaces and detailed analysis of biological samples. 

Our contributions are summarized as follows:
\begin{itemize}
    \item To the best of our knowledge, we are the first to propose a method for textured 3D reconstruction of microstructures using only a commercial WLI and an OM, without costly hardware modifications needed in previous methods.
    \item We propose a novel pipeline that includes a two-step data association method and a neural implicit representation with a residual term to fuse multi-modal data, enabling accurate geometric reconstruction and natural surface color acquisition.
    \item We evaluate the proposed method on a dataset of multi-view WLI and OM data of various real-world microscale samples, as well as on a synthetic dataset, demonstrating the effectiveness and practical value of our approach.
\end{itemize}

\section{Related Work}
\label{sec:related_work}
\textbf{Multi-Modal 3D Reconstruction.}
Multi-modal 3D reconstruction methods enhance the accuracy and speed or provide richer 3D information by combining data from different sensors and leveraging their unique advantages.
Beyond commonly used RGB cameras, these methods also incorporate sensors such as depth cameras~\cite{kinectfusion, voxel_RGBD, RGBD_recon, RGB_TOF}, Lidar~\cite{RGB_Lidar}, infrared (IR)~\cite{IR_RGBD} and thermal imaging~\cite{thermal_RGB, thermal_RGB2}. In our work, OM and WLI can be approximated as a combination of an RGB camera and a depth sensor.
Conventional RGBD reconstruction methods use depth information for model reconstruction~\cite{kinectfusion, voxel_RGBD} and RGB for subsequent appearance reconstruction steps to texture the 3D model~\cite{color_map, mvs_texturing}.
Recent methods~\cite{RGBD_recon, RGB_TOF, nice_slam} use neural networks to implicitly represent 3D models, enabling the fusion of multi-sensor information into the 3D model through an end-to-end optimization process.
Our work shifts the focus from macroscale to microscale objects. Currently, research on 3D reconstruction of microscale surfaces is relatively limited, with existing studies~\cite{3DSEM_survey, MVN-AFM} typically relying on a single instrument. In microscale research, a range of available instruments exhibit entirely different characteristics. Our work is the first attempt to fuse WLI and OM data, exploring the feasibility of these multi-modal 3D reconstruction methods in the microscale domain.

\noindent\textbf{Neural Implicit Representation.}
Neural implicit representations are widely applied to various tasks. DeepSDF~\cite{deepsdf} employs neural networks to represent 3D shapes as signed distance functions. NeRF~\cite{nerf} and its subsequent works~\cite{nerf_w, pixel_nerf, mip_nerf360} model scenes as radiance fields, allowing for realistic novel view synthesis. 
Some approaches~\cite{nerfactor, intrinsicnerf} have implemented the decomposition of lighting and color attributes within a scene, thereby supporting more flexible applications.
Further developments, such as NeuS~\cite{neus} and VolSDF~\cite{volsdf}, focus on accurate surface reconstruction, enhancing the fidelity of 3D models. Other works ~\cite{neural_dynamic, nerfies} extend these representations to dynamic scenes, enabling the capture of temporal changes. Moreover, neural implicit representations have proven effective in handling data from other modalities, such as CT~\cite{CT_neural} and MRI~\cite{MRI_Nerp}.
Our work further demonstrates the effectiveness of neural implicit representations in fusing multi-view WLI and OM data. Our method captures the fine surface geometric features and intrinsic colors of microstructures, which are essential for microscale research and applications.

\section{Method}
\label{sec:method}

\begin{figure*}[h] 
    \centering
    \includegraphics[width=\linewidth]{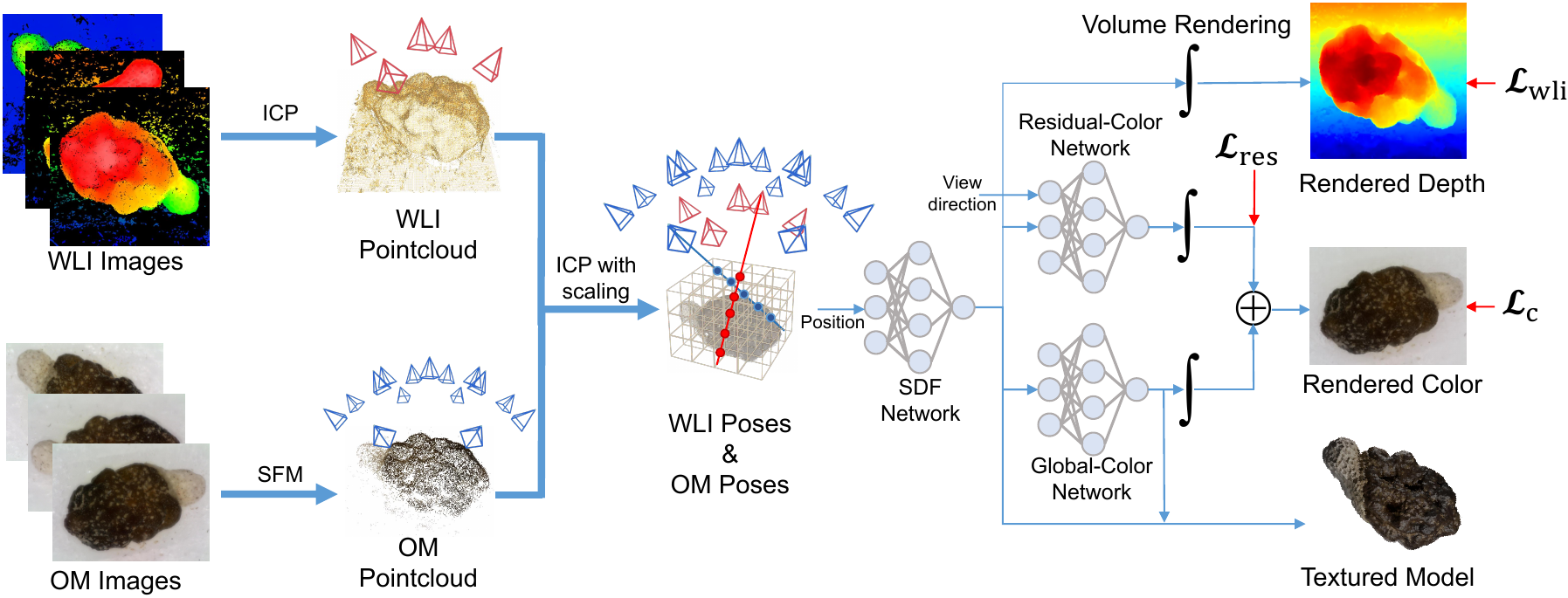}
    \caption{ \textbf{System pipeline of OpticFusion.} Our method takes multi-view WLI and OM images of a microscale sample as input. Through a two-step data association process, we compute the pose of each WLI and OM image in the same absolute-scale coordinate system. Using differentiable volume rendering techniques, we train the neural implicit representation of the microstructure's geometry and view-independent color under the supervision of the multi-modal data. The result is a textured 3D surface model of the microscale sample.}
    \label{fig: system_pipeline}
    \vspace{-1.5em}
\end{figure*}

In this section, we introduce the characteristics of WLI data in \cref{sec:preliminary}. Here, we design the first pipeline for textured surface reconstruction of microscale samples using WLI and OM multi-modal data, as shown in \cref{fig: system_pipeline}. Specifically, given multi-view WLI and OM data of a microscale sample as input, we obtain the camera poses for these two sets of multi-modal data within an absolute-scale coordinate system using a two-step data association process (\cref{sec:data_alignment}). In \cref{sec:3d_recon}, we use both conventional reconstruction and texture mapping workflows (\cref{sec:conventional_recon}), as well as neural implicit surface reconstruction workflows (\cref{sec:neural_recon}), to fuse the multi-modal data into a textured 3D model of the sample. Finally, we present the implementation details of the neural implicit surface reconstruction in \cref{sec:optimization}.

\subsection{Preliminary: The Characteristics of WLI Data}
\label{sec:preliminary}
The WLI utilizes interference from a broad-spectrum light source to achieve high-precision 3D measurements of microscale sample surfaces. The output of a commercial WLI instrument is a height map with an accurate scale, where each pixel's measurement precision can reach sub-nanometer levels. With the known XY scanning range corresponding to the objective lens used in WLI, the absolute coordinates of each pixel can be calculated. 
Thus, each WLI image can be represented as a point cloud or, as discussed in subsequent sections, a depth map from an orthographic camera. 
Although WLI offers high measurement accuracy, it encounters limitations when measuring smooth micro surfaces with steep angles, resulting in incomplete data with significant voids (more details in Sec.~\ref{sec:evaluation_sim_dataset}). Additionally, WLI instruments are expensive and relatively slow when scanning areas with large height variations. Consequently, in real experiments, it is challenging to obtain a substantial number of WLI scans from multiple viewpoints for a single sample. 
Therefore, the multi-view WLI data used in our experiments can be described as a set of high-precision depth maps with limited views and containing data voids.

\subsection{Multi-Modal Data Association}
\label{sec:data_alignment}
First, we need to obtain the poses of virtual orthographic cameras corresponding to the multi-view WLI images, as well as the camera poses of the multi-view OM images.
Most multi-modal 3D reconstruction methods involve multiple sensors with fixed relative rigid-body transformations and synchronized, aligned data. 
For example, an RGBD camera provides aligned color and depth images, or different sensors fixed on the same platform with unchanging relative poses. 
However, in our approach, the acquisitions of WLI and OM images are completely independent.
Moreover, WLI and OM images have different resolutions, different numbers of inputs, and different imaging models, making it challenging to directly establish data associations between the raw WLI and OM images. To address this issue, we propose a two-step data association method.

In the first step, we separately calculate the relative camera poses among multiple WLI images and among multiple OM images. Multi-view WLI images are converted into 3D point clouds, and then the Iterative Closest Point (ICP) algorithm~\cite{ICP} is used to align these point clouds within a unified coordinate system. This process calculates the relative poses of the virtual orthographic cameras of the WLI images and produces a merged WLI point cloud, denoted as $P_{\text{wli}}$. For multi-view OM images, we use traditional structure from motion (SFM) ~\cite{Agisoft_Metashape} to compute the camera poses and generate a sparse point cloud of the microstructure. We then filter out low-confidence points, retaining only high-confidence spatial points, which results in the OM point cloud, denoted as $P_{\text{om}}$.
In the second step, we calculate the relative poses between the WLI and OM images by aligning two intermediate point clouds. Since WLI data contains absolute-scale information while the OM reconstruction result has scale ambiguity, we employ the ICP algorithm with scale transformation to align $P_{\text{om}}$ to the $P_{\text{wli}}$ coordinate system. Finally, this alignment allows us to obtain the camera poses for each WLI and OM image within the same absolute-scale coordinate system.

\subsection{Multi-Modal 3D Reconstruction}
\label{sec:3d_recon}
\subsubsection{Conventional Solution}
\label{sec:conventional_recon}
A conventional method for fusing multi-modal data for 3D reconstruction involves two sequential steps: surface reconstruction and texture mapping. Here, we utilized this conventional approach to fuse WLI and OM data, reconstructing textured 3D models of microscale samples. First, we used the Poisson surface reconstruction method ~\cite{screened_poisson} to convert the fused point cloud obtained from the previous steps into a surface mesh of the microstructure. Then, we applied the MVS-Texturing~\cite{mvs_texturing} method to map the OM images onto the mesh, resulting in a textured 3D model.

\subsubsection{Neural Implicit Multi-Modal 3D Reconstruction with View-Independent Color}
\label{sec:neural_recon}
Our multi-modal neural implicit representation consists of three networks: an SDF network, a global color network, and a residual color network. 
The SDF network converts input coordinate $\mathbf{x}$ into the SDF value at that spatial location and a geometry feature vector $\mathbf{f}$. We follow the approaches of NeuS~\cite{neus} in the design of this part.
In the original NeuS, $\mathbf{f}$ is immediately input into a color network along with the view direction $\mathbf{v}$ and normal $\mathbf{n}$ to map the spatial point to its color value in the specified view direction.
Importantly, unlike rendering applications for macroscale objects, research in the microscale domain does not require the synthesis of novel views. Instead, it focuses on the geometry of the sample and its intrinsic colors. Therefore, we do not want the reconstructed surface color of the 3D model including surface highlights, due to lighting or changes in viewing direction.
Inspired by the color decomposition strategies in Intrinsic-NeRF~\cite{intrinsicnerf} and Color-NeuS~\cite{color_neus}, we modified NeuS to decouple the view-dependent color components and eliminate the specular highlights from the object surface in the OM images. 
The original color network is further decomposed into a view-dependent residual color network $\mathcal{R}$ and a view-independent global color network $\mathcal{G}$. 
\begin{equation}
c_{\mathrm{res}} = \mathcal{R}(\mathbf{x}, \mathbf{n}, \mathbf{f}, \mathbf{v}),
\end{equation}
\begin{equation}
c_{\mathrm{global}} = \mathcal{G}(\mathbf{x}, \mathbf{n}, \mathbf{f}).
\end{equation}
The outputs of the two networks are added together to get the complete color $c=c_{\mathrm{res}}+c_{\mathrm{global}}$. We follow the volume rendering of NeuS~\cite{neus} to accumulate the output into a rendered color $C$. When we generate the mesh with the marching cube algorithm, only the global color network is used to infer the vertex colors of the mesh to get the final textured 3D model of the microscale sample.

\subsection{Optimization}
\label{sec:optimization}
To optimize the implicit representation networks for the geometry and color of microstructures, we minimize a series of losses, including color loss, residual loss, WLI loss, and SDF regularization loss, supervised by the posed WLI and OM images.
We randomly sample pixels on both WLI and OM images and generate sampling rays based on the corresponding camera projection model and pose. In our experiments, WLI and OM image pixels have the same probability of being sampled.
We denoted $m$ as the batch size and $n$ as the number of sample points.
First, the color loss is defined as the difference between the rendered color $C$ along the ray and the pixel color $\widehat{C}$ in the OM image:
\begin{equation}
\mathcal{L}_{\mathrm{c}} = \frac{1}{m}\sum_{i}
{\left\lVert \widehat{C_i} - C_i  \right\rVert}_2^2.
\end{equation}
Second, the residual loss aims to minimize the output of the residual color network:
\begin{equation}
\mathcal{L}_{\mathrm{res}} = \frac{1}{m} \sum_{i} 
{\left\lVert C_{\mathrm{res}_i}  \right\rVert}_2^2.
\end{equation}
This guides the view-independent global color network in capturing the primary color information of the OM image. In contrast, the residual network only represents other view-dependent components, such as specular highlights on the object's surface.  

The WLI loss is the difference between the depth $D$ along the ray direction and the WLI depth $\widehat{D}$:
\begin{equation}
\mathcal{L}_{\mathrm{wli}} = \frac{1}{m}\sum_{i} 
{\left\lVert \widehat{D}_i - D_i  \right\rVert}_2^2.
\end{equation}

Finally, a commonly used regularization term $\mathcal{L}_{\mathrm{reg}}$~\cite{SDF_Regular}, is applied to constrain the network's SDF output:
\begin{equation}
\mathcal{L}_{\mathrm{reg}} = \frac{1}{mn} \sum_{i,j} \left( \|
\mathbf{n}_{i,j}\| - 1\right)^2.
\end{equation}

However, unlike RGBD images where each pixel has both color and depth information, each sampled point on WLI and OM images only has depth or color information, respectively. Therefore, when sampling a pixel on an OM image, the loss function is \(\mathcal{L} = \lambda_{\mathrm{c}} \mathcal{L}_{\mathrm{c}} + \lambda_{\mathrm{res}} \mathcal{L}_{\mathrm{res}} + \lambda_{\mathrm{reg}} \mathcal{L}_{\mathrm{reg}}\). When sampling a pixel on a WLI image, the loss function is \(\mathcal{L} = \lambda_{\mathrm{wli}} \mathcal{L}_{\mathrm{wli}} + \lambda_{\mathrm{reg}} \mathcal{L}_{\mathrm{reg}}\).
\section{Experiments}
\label{sec:experiments}
\begin{figure*}[h] 
    \centering
    \includegraphics[width=\linewidth]{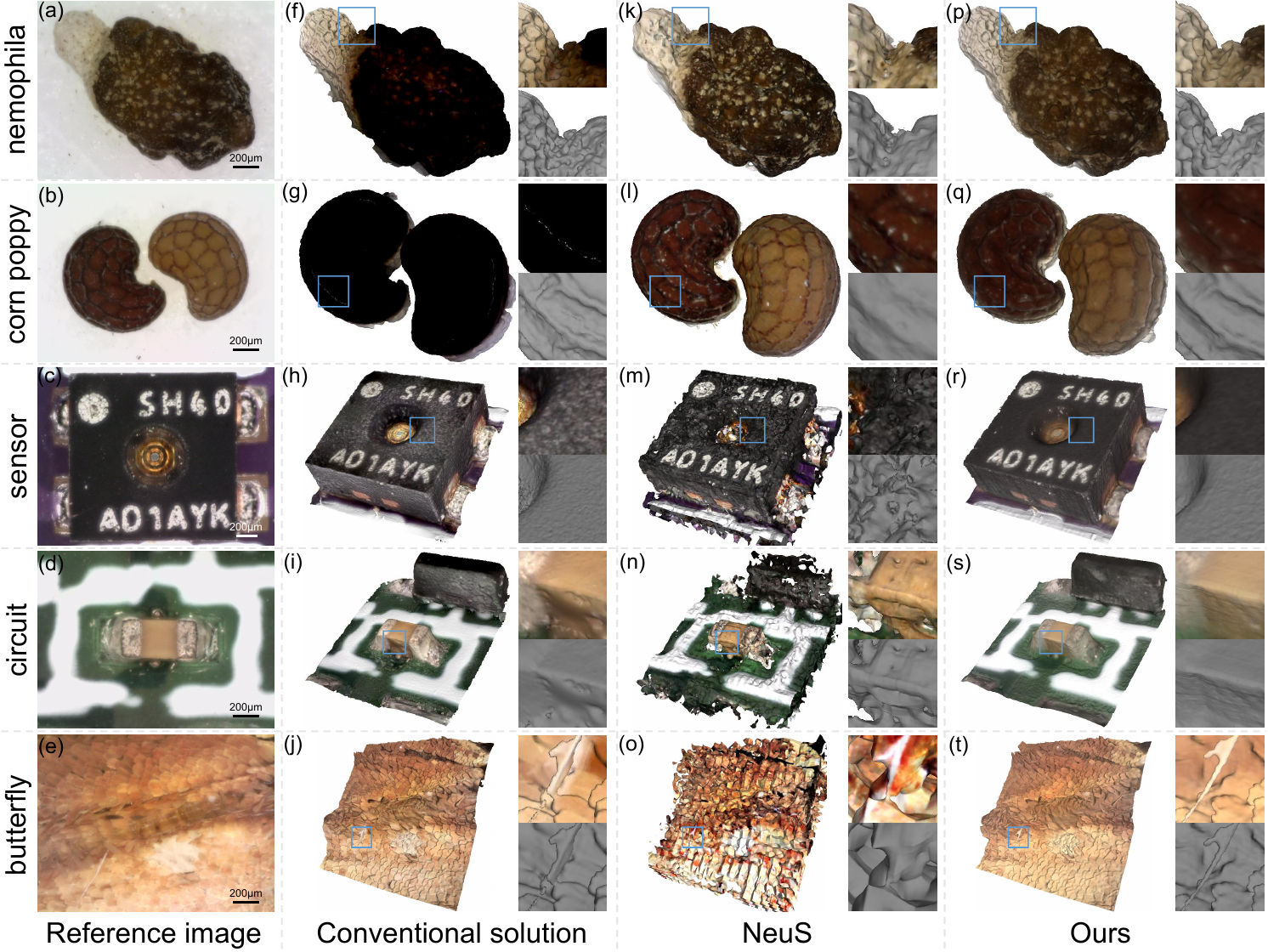}
    \caption{ \textbf{Qualitative reconstruction results on our WLI-OM real-world dataset.} Compared with other methods, our reconstruction results have more accurate surface geometry and natural texture colors. Please see more visualizations in the supplementary material.}
    \label{fig: real_results}
    \vspace{-1em}
\end{figure*}

\subsection{Experimental Setup}
In our experiments, we executed the OpticFusion code on a workstation computer equipped with an AMD 5950X CPU and an Nvidia RTX4090 GPU.
Our multi-modal neural implicit surface reconstruction is built upon the existing PyTorch~\cite{pytorch} implementation~\cite{instant-nsr-pl} of hash encoding~\cite{instant_ngp} and NeuS~\cite{neus}.
The loss weights for our experiments are set as follows: $\lambda_{\mathrm{c}} = 1.0$, $\lambda_{\mathrm{res}} = 0.001$, $\lambda_{\mathrm{wli}} = 1.0$, and $\lambda_{\mathrm{reg}} = 0.1$.
We used the Adam optimizer~\cite{adam} with a learning rate of 0.01 to train the network, conducting 20,000 iterations. Each iteration involved randomly selecting 256 rays and sampling 1024 points per ray. Please refer to the supplementary material for more implementation details.

\subsection{Evaluation on Real World Dataset}
\noindent\textbf{WLI-OM Dataset.}
We first constructed a dataset comprising multi-view OM images and WLI scans of real-world microscale samples. This dataset includes five sequences collected from typical microstructures studied in microscale research, such as flower seeds, a butterfly wing, a circuit board, and an electronic component. The surface geometric features of these samples range in size from tens to hundreds of micrometers.
The OM images were captured by a commercial digital microscope, Dino-Lite AM7915MZT. 
Each sample was imaged by OM at tilt angles of 15$^{\circ}$, 30$^{\circ}$, and 45$^{\circ}$, with a photo taken every 30$^{\circ}$ around the z-axis, plus one top-down view.
This process resulted in 37 OM images per sample, each with a resolution of 1296 $\times$ 972.
For the multi-view WLI data, we used a commercial WLI instrument, ZYGO NewView 8200. Since the WLI can only move along the z-axis and cannot tilt, we placed the samples on a 30$^{\circ}$ tilt turntable and performed scans every 90$^{\circ}$. Together with a scan without tilt, we obtained 5 WLI images per sample, each with a resolution of 1024 $\times$ 1024.

\noindent\textbf{Results Analysis.}
We applied three methods to the WLI-OM dataset: the conventional solution described previously, the NeuS method, and our OpticFusion method. The results are shown in \cref{fig: real_results}.
While WLI offers very high measurement accuracy, it often results in significant voids in the data, with missing measurements on some parts of the sample surface. Consequently, Poisson reconstruction in the conventional solution causes unsatisfactory geometry, as seen in the apparent surface defects in \cref{fig: real_results}\textcolor{red}g and \ref{fig: real_results}\textcolor{red}i and the discontinuities on the butterfly's tiny surface burr in \cref{fig: real_results}\textcolor{red}j. Additionally, the conventional texture mapping method can sometimes fail, resulting in inconsistencies with the OM images.
For NeuS, we used only OM images as input. However, due to the limited number and angles of the OM images, the reconstructed geometry is poor in some texture-less areas of the microstructure surface. Furthermore, because the optical microscope cannot capture very subtle shape variations on the sample surface, some surface details are lost in the reconstructed results, such as the white areas in \cref{fig: real_results}\textcolor{red}k and the depressions on the surface in \cref{fig: real_results}\textcolor{red}l.
In contrast, our OpticFusion method performs better in fusing multi-view WLI and OM data into a textured 3D model. The multi-view OM images help fill in the areas where WLI data is missing, such as the surfaces in \cref{fig: real_results}\textcolor{red}q and \ref{fig: real_results}\textcolor{red}s and the complete tiny burr on the butterfly wing in \cref{fig: real_results}\textcolor{red}t. The WLI data provides accurate constraints for surface reconstruction, preventing the surface collapse in texture-less regions, as observed in NeuS. The few input OM images are enough to texture the surface with natural colors.
These results demonstrate that our OpticFusion method effectively combines the advantages of WLI and OM data, achieving high-fidelity geometry and texture reconstruction even in challenging microscopic scenarios, ensuring robustness and detail retention across varying surface features.

\subsection{Application}
\label{sec:application}
\begin{figure}[h] 
    \centering
    \includegraphics[width=\linewidth]{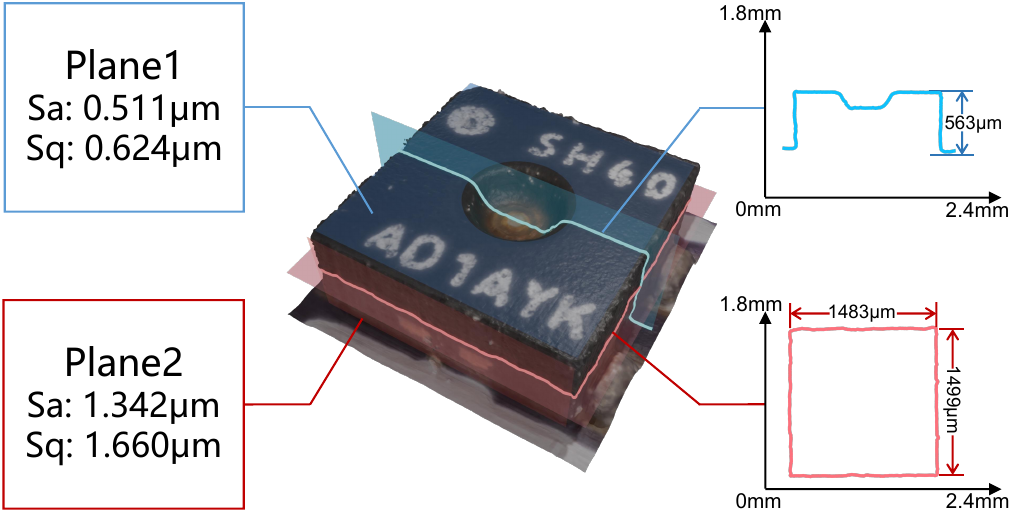}
    \caption{ \textbf{Roughness measurement of microscale surfaces.} We calculate the roughness of two surfaces and obtain cross-sections in two directions.}
    \label{fig: roughness_analysis}
    \vspace{-1em}
\end{figure}

Our method has broad applications in microscale research, as demonstrated by the following examples. Surface roughness analysis of microstructures is a common standard for evaluating the precision of micro-nanoscale processing~\cite{li2018calibration, zhang2019manufacturing, lucca2020ultra}. By obtaining a complete 3D model of the microscale sample with our method, we can easily analyze the surface roughness of any sample region. Here, we use two roughness parameters, the arithmetical mean height of the surface (Sa) and the root mean square height of the surface (Sq), with calculation details provided in the supplementary material. The roughness analysis results on different surfaces of a microsensor are shown in \cref{fig: roughness_analysis}.

\begin{figure}[h] 
    \centering
    \includegraphics[width=\linewidth]{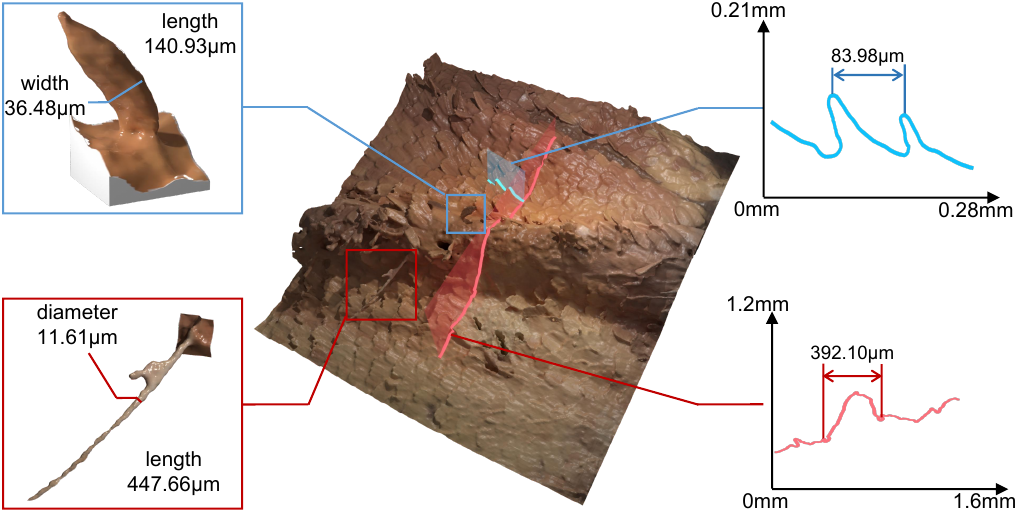}
    \caption{ \textbf{Surface detail analysis of a biological sample.} We isolate separate 3D models of a single scale and a tiny burr on the butterfly's wing. Additionally, we provide cross-sections of multiple continuous scales and veins.}
    \label{fig: detail_analysis}
    \vspace{-1em}
\end{figure}

Our reconstruction results are also highly beneficial for analyzing biological surface details~\cite{spider_WLI}. Cross-sectional images of the butterfly wing in different directions clearly show the distribution, spacing, and size of surface scales, as well as the thickness and height of wing veins (\cref{fig: detail_analysis}). Researchers can even select and analyze specific structures on the biological surface, such as an individual scale or a tiny burr. The reconstructed complete 3D models with natural colors provide a more intuitive and detailed perspective for researchers in the microscopic field, facilitating further analysis and understanding of microstructures.

\subsection{Evaluation on Synthetic Dataset}
\label{sec:evaluation_sim_dataset}
\begin{figure}[h] 
    \centering
    \includegraphics[width=\linewidth]{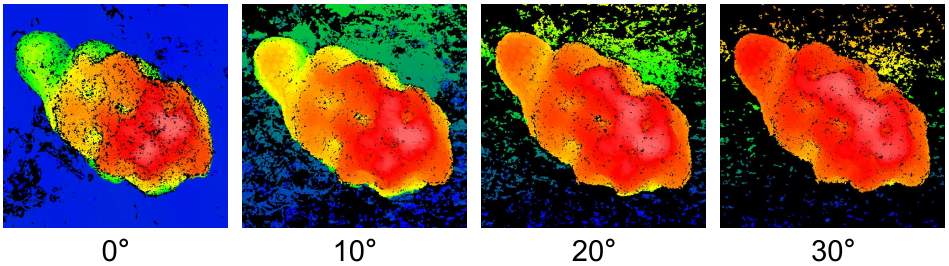}
    \caption{ \textbf{WLI scanning results at different angles.} As the scanning angle increases, the proportion of valid measurement data from WLI on the specular surface (bottom plane) gradually decreases. In contrast, the rough scattering surface (seed surface) does not change significantly.}
    \label{fig: multi_angle}
    \vspace{-1em}
\end{figure}

\begin{figure}[h] 
    \centering
    \includegraphics[width=0.98\linewidth]{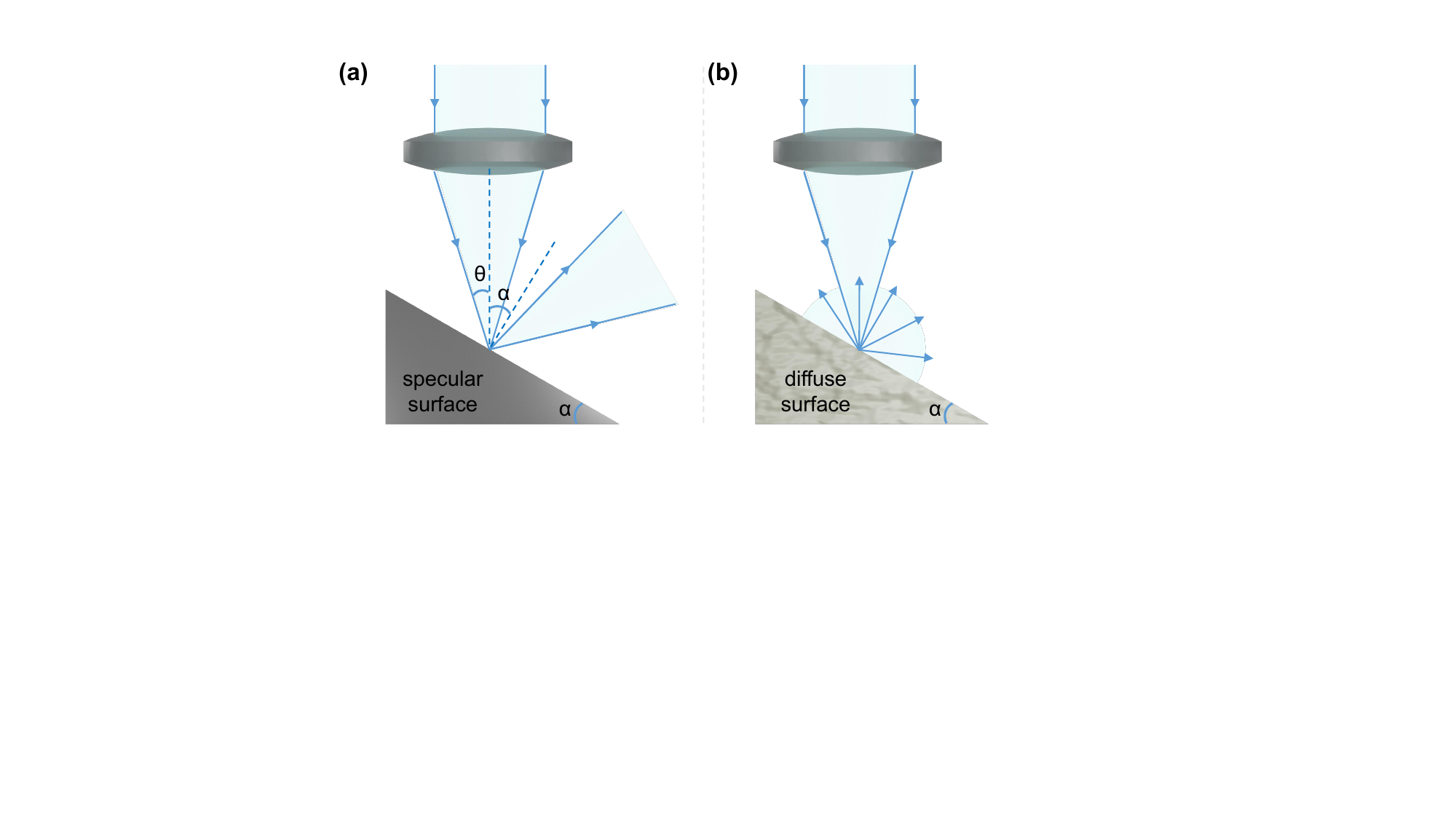}
    \caption{ \textbf{WLI measurements of two different types of surfaces at tilt angles exceeding the numerical aperture (NA) limit.} \textbf{(a)} on specular surface. \textbf{(b)} on diffuse surface.}
    \label{fig: NA_limit}
    \vspace{-1em}
\end{figure}

\noindent\textbf{Construction of Synthetic Dataset.}
In real experimental settings, obtaining the ground truth 3D model of microscale samples is challenging, as WLI is among the most precise instruments for 3D measurements in microstructures. To quantitatively evaluate our method, we constructed a set of synthetic multimodal data based on the configurations of our WLI-OM dataset.
We utilized 3D models from the NeRF Synthetic dataset~\cite{nerf} in our synthetic dataset. We rendered RGB images with known poses in Blender~\cite{Blender}, matching the number, tilt angles, and resolution of OM images in the WLI-OM dataset.
The main challenge is simulating multi-view WLI data. WLI data can be regarded as noise-free depth maps acquired by an orthographic camera since WLI’s sub-nanometer measurement error is negligible compared to the geometric features of microscale samples.
A key characteristic of WLI is that the numerical aperture (NA) of the objective lens limits the measurable surface tilt angles~\cite{NA_Limit1, NA_Limit2}, leading to voids in the scan data (\cref{fig: multi_angle}). 
As shown in \cref{fig: NA_limit}\textcolor{red}a, when the tilt angle $\alpha$ of a specular micro surface exceeds $\theta = arcsin(\mathrm{NA})$, the incident light cannot be reflected back to the WLI objective lens, resulting in no observable interference pattern on the CCD camera. Thus, many areas on the smooth substrate of the sample at large tilt angles lack measurement values in \cref{fig: multi_angle}. Conversely, rough micro surfaces can exceed the NA limitation because the diffuse and back-scattered light from the rough tilted surface can be captured by the objective lens (\cref{fig: NA_limit}\textcolor{red}b). High dynamic range measurement techniques capture interference fringes, providing measurements on the rough seed surface in \cref{fig: multi_angle}.
In our synthetic data, we approximate this relationship between the proportions of the voids in WLI data and the surface tilt angle and reflectance by selectively removing depth values from the original depth maps (more details are included in the supplementary).

\begin{table*}[h]
  \centering
  \small
  \begin{tabular}{c c c c c c c c c c}
    \toprule
    Chamfer distance $\downarrow$ &lego&drums&hotdog&ficus&chair&mic&ship&average\\
    \midrule
    Conventional solution & 0.511 & 1.248 & 0.341 &	0.178 &	1.188 &	1.858 &	0.227 & 0.793\\
    Ours w/o WLI supervision & 0.160	& 0.034 & 1.067 & 0.021 & 0.193 & 0.026 & 2.265 & 0.538\\
    Ours w/o OM supervision & 0.148 & 0.491 & \textbf{0.053} & 0.049 & 0.614 & 0.534 & 2.855 & 0.678\\
    Ours w/o residual term & 0.038 & 0.033 & 0.065 & \textbf{0.017} & 0.071 & 0.016 & \textbf{0.069} & 0.044\\
    Ours & \textbf{0.031} & \textbf{0.030} & 0.071 & 0.018 & \textbf{0.064} & \textbf{0.015} & 0.070 & \textbf{0.043}\\
    \bottomrule
  \end{tabular}
  \vspace{-0.5em}
  \caption{ \textbf{Quantitative comparison of reconstruction results on the synthetic multi-view WLI and OM dataset.} We use chamfer distance as our evaluation metric. Our method performs better than other methods. Moreover, the residual item in our method has no noticeable impact on the reconstruction quality.}
  \vspace{-1.5em}
  \label{tab:sim_chamfer}
\end{table*}

\noindent\textbf{Results Analysis.}
Here, we use Chamfer Distance to evaluate the reconstruction quality of different methods: conventional solution (Poisson reconstruction of the WLI point cloud), training a neural implicit representation (NeuS) with either multi-view OM data or WLI data separately, using both WLI and OM for supervision without the residual term, and our complete reconstruction method, OpticFusion. The results are shown in \cref{tab:sim_chamfer}.
First, the Poisson reconstruction of the incomplete WLI point cloud yields relatively poor results. Second, The results demonstrate that multimodal reconstruction using both WLI and OM data significantly outperforms reconstruction using either data type alone, consistent with our observations in the real data experiment. These results indicate that fusing WLI and OM data through neural implicit representation is an effective approach for reconstructing complex 3D structures at the microscopic level.
Finally, for the residual term used in our OpticFusion, we intend for it to improve only the texture color of the final model without negatively impacting the geometric reconstruction. As expected, the residual term does not significantly affect the reconstruction error. Please refer to our supplementary material for more visualizations.

\subsection{Ablation Study}

\begin{figure}[h] 
    \centering
    \includegraphics[width=\linewidth]{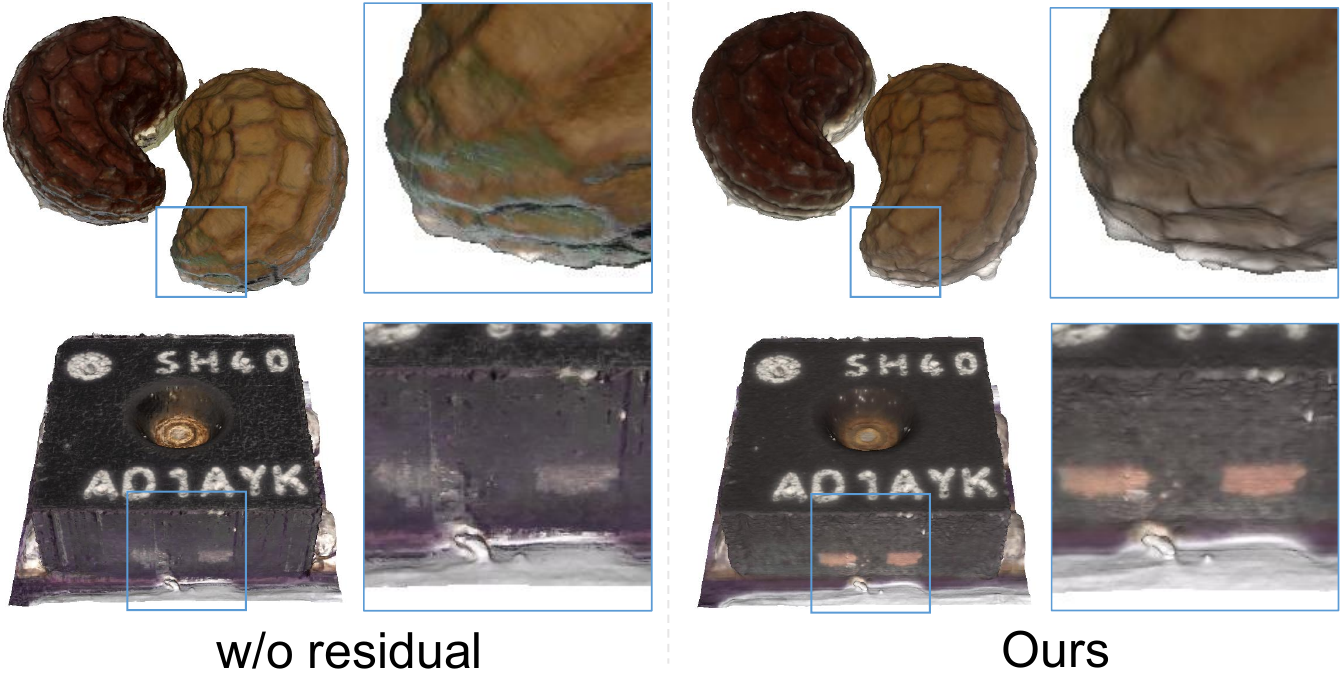}
    \caption{ \textbf{Effect of the residual term.} }
    \label{fig: residual_ablation}
    \vspace{-1em}
\end{figure}

As shown in \cref{fig: residual_ablation}, we demonstrate the necessity and effectiveness of introducing a residual term during the generation of surface textures. A common method for calculating the color of vertices extracted in neural implicit surface representation is to use the direction opposite to the surface normal as the view direction and then infer the color using the color network. However, for microscale samples, the limited number and angles of optical microscope observations make it difficult to obtain comprehensive views of each point from various directions. Additionally, highlights at particular angles on the sample surface may obscure the original color of the sample. Therefore, in our real experimental data, we clearly observe that this common texture generation method is insufficient, leading to some strange colors appearing on our model. By introducing the residual term, the color of each vertex is obtained by inputting the coordinates of the point into the view-independent global color network, resulting in a more natural surface color.

\begin{figure}[h] 
    \centering
    \includegraphics[width=\linewidth]{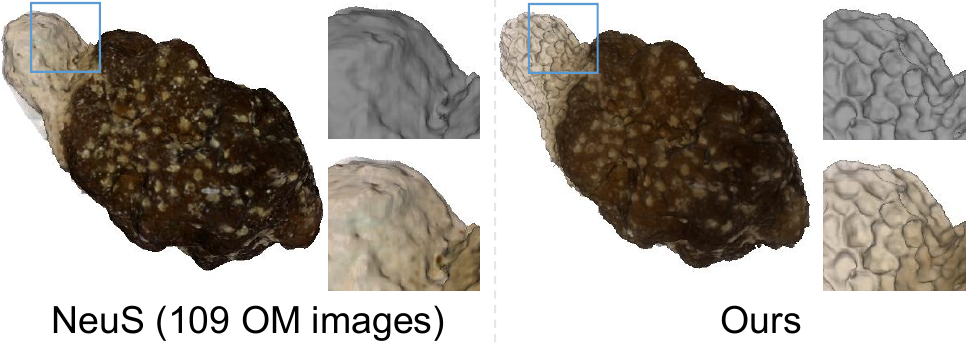}
    \caption{ \textbf{Reconstruction result with more OM images.} }
    \label{fig: more_OM}
    \vspace{-1.9em}
\end{figure}

We further explored the performance of dense reconstruction relying solely on OM images with a sufficient number of views. Here, we collected 109 OM images of one sample from the WLI-OM dataset, three times the number used in our previous experiments. As shown in \cref{fig: more_OM}, the OM images still fail to reconstruct certain detailed grooves on the sample's surface. This is because the geometric information in these areas is not represented as texture information in the OM images, and thus cannot be obtained through multi-view reconstruction. On the other hand, the ability of WLI to measure surface topography is independent of the sample's surface texture and mainly depends on whether the sample can reflect the incident white light back into the objective lens. This further demonstrates the necessity and effectiveness of fusing WLI and OM data.
\section{Conclusion}
\label{sec:conclusion}
We propose a novel method for neural implicit surface reconstruction of microstructures using WLI and OM multi-modal data to obtain textured 3D models of microscale samples. To get the camera poses of independently collected WLI and OM data, we employ a two-step data association method. We then use multi-modal data as supervision to optimize the neural implicit representation, effectively combining the advantages of WLI and OM data to achieve precise reconstruction of microscale surface geometry. 
We use a color decomposition strategy that introduces a view-dependent residual term into the implicit representation and extracts the view-independent color component from OM images, resulting in more natural texture colors. Our experiments demonstrate OpticFusion's capability to reconstruct high-quality textured 3D models of various microscale samples and its practical value in microscopic research.

{\small\paragraph{Acknowledgements.}
We thank Yuan-Liu Chen for his assistance in the WLI experiment. This work was partially supported by NSF of China (No. 61932003).}

{
    \small
    \bibliographystyle{ieeenat_fullname}
    \bibliography{main}
}
\appendix
\clearpage
\maketitlesupplementary

\setcounter{table}{0}
\setcounter{figure}{0}
\setcounter{equation}{0}
\renewcommand{\thetable}{\thesection\arabic{table}}
\renewcommand{\thefigure}{\thesection\arabic{figure}}
\renewcommand{\theequation}{\thesection\arabic{equation}}

\section{Network Details}
\label{sec:network}
The input spatial coordinates are initially encoded using the multi-resolution hash technique~\cite{instant_ngp}. This encoding involves 16 levels of hash feature grids, each with feature dimensions of 2. 
The coarsest grid starts with a resolution of 16, and each subsequent level increases by a scale factor of 1.447.
The SDF network consists of an MLP with one hidden layer of size 64, utilizing ReLU activation. For the global and residual color networks, each is structured as an MLP with two hidden layers of size 64, also employing ReLU activation. The output from these color networks is normalized to a range of 0 to 1 using the sigmoid function.

\section{Computation of Roughness Parameters}
\label{sec:roughtness}
As demonstrated in Sec. \ref{sec:application} of our main paper, we provide two roughness parameters for microscale surfaces: the arithmetical mean height of the surface (Sa) and the root mean square height of the surface (Sq).
For a selected surface area, we first fit the surface to a plane and then rotate and translate it to align with the XY plane. Then, on this transformed plane, $N$ points are uniformly sampled at equal intervals along the XY coordinates, and the Z-axis heights $z$ of these points are recorded. The arithmetic mean of the heights $\bar{z}$ is calculated as follows:
\begin{equation}
\bar{z} = \frac{1}{N} \sum_{i=1}^{N} z_i.
\end{equation}
Sa is defined as the mean absolute height difference between the surface points and the mean plane:
\begin{equation}
\mathrm{Sa} = \frac{1}{N} \sum_{i=1}^{N} |z_i - \bar{z}|.
\end{equation}
Sq represents the standard deviation of the point heights:
\begin{equation}
\mathrm{Sq} = \sqrt{\frac{1}{N} \sum_{i=1}^{N} (z_i - \bar{z})^2}.
\end{equation}

\section{Generation of Synthetic WLI Data}
\label{sec:wli_generation}
\begin{figure}[h] 
    \centering
    \includegraphics[width=\linewidth]{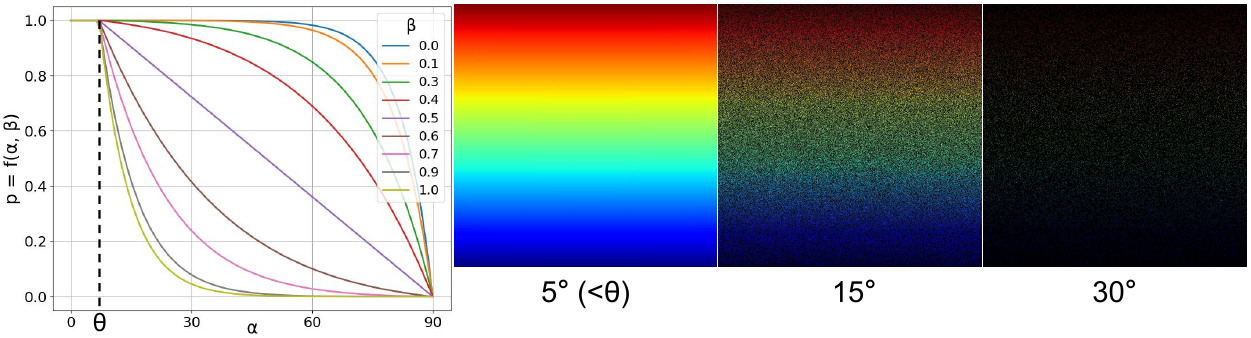}
    \vspace{-1em}
    \caption{ \textbf{ The curve of the function $f$, and synthetic WLI data for planes with different tilt angles.}}
    \label{fig: WLI_curve}
    \vspace{-2em}
\end{figure}

Here, We demonstrate how we generate synthetic WLI data in our synthetic experiment. Real WLI data has high measurement accuracy and can be regarded as noise-free depth maps in the simulated environment. 
Moreover, WLI data often contains voids because the CCD camera fails to capture interference fringes with sufficient contrast on some surfaces. 
This issue typically arises when the sample's tilt angle $\alpha$ (the angle between the surface normal and the Z-axis) is too large, preventing the white light from reflecting back into the objective lens.
As discussed in Sec. \ref{sec:evaluation_sim_dataset}, for a perfectly specular micro surface, this tilt angle limit can be calculated from the numerical aperture (NA) of the objective lens, NA is 0.13 in our real experiments. On the other hand, rough surfaces with larger tilt angles can still have measurements.
We summarize the characteristics of WLI data generation as follows:
\begin{itemize}
    \item When the tilt angle is less than \(\theta=\arcsin(\text{NA})\), measurements can be obtained on any surface.
    \item When the tilt angle exceeds \(\theta\), specular surfaces are less likely to yield measurements, while diffuse surfaces are more likely to do so.
\end{itemize}
Our objective is to ensure that the synthetic WLI data replicates these characteristics of real WLI data. To achieve this, we render a depth map \(d\), normal map \(n\), diffuse color \(c_d\), and specular color \(c_s\) from each viewpoint in Blender. Using \(n\) and the view direction, we calculate the surface tilt angle \( \alpha \) for each pixel. We then define \(\beta = \frac{c_s}{c_d + c_s}\) as the reflectance of the model surface. Based on the first rule, we design a piecewise function \(f\) to calculate the probability \(p\) that the depth value of each pixel is valid:
\begin{equation}
p = f(\alpha, \beta) =
\begin{cases}
1, & \text{for } \alpha < \theta, \\
g, & \text{for } \alpha \geq \theta.
\end{cases}
\end{equation}
Following the second rule, we define the function \(g\). We set two intermediate variables \(x = \frac{\alpha - \theta}{90 - \theta}\), and \(t = 2\beta - 1\):
\begin{equation}
g(x, t) =
\begin{cases}
1 - e^{10(1-x)t} \cdot x, & \text{for } t < 0, \\
e^{-10xt} \cdot (1-x), & \text{for } t \geq 0. 
\end{cases}
\end{equation}
As illustrated in \cref{fig: WLI_curve}, the $f$ curve closely aligns the characteristics of real WLI data. 
We then randomly remove values in $d$ according to the calculated $p$ of each pixel and then obtain synthetic WLI data that contain voids.

\begin{figure*}[h] 
    \centering
    \includegraphics[width=\linewidth]{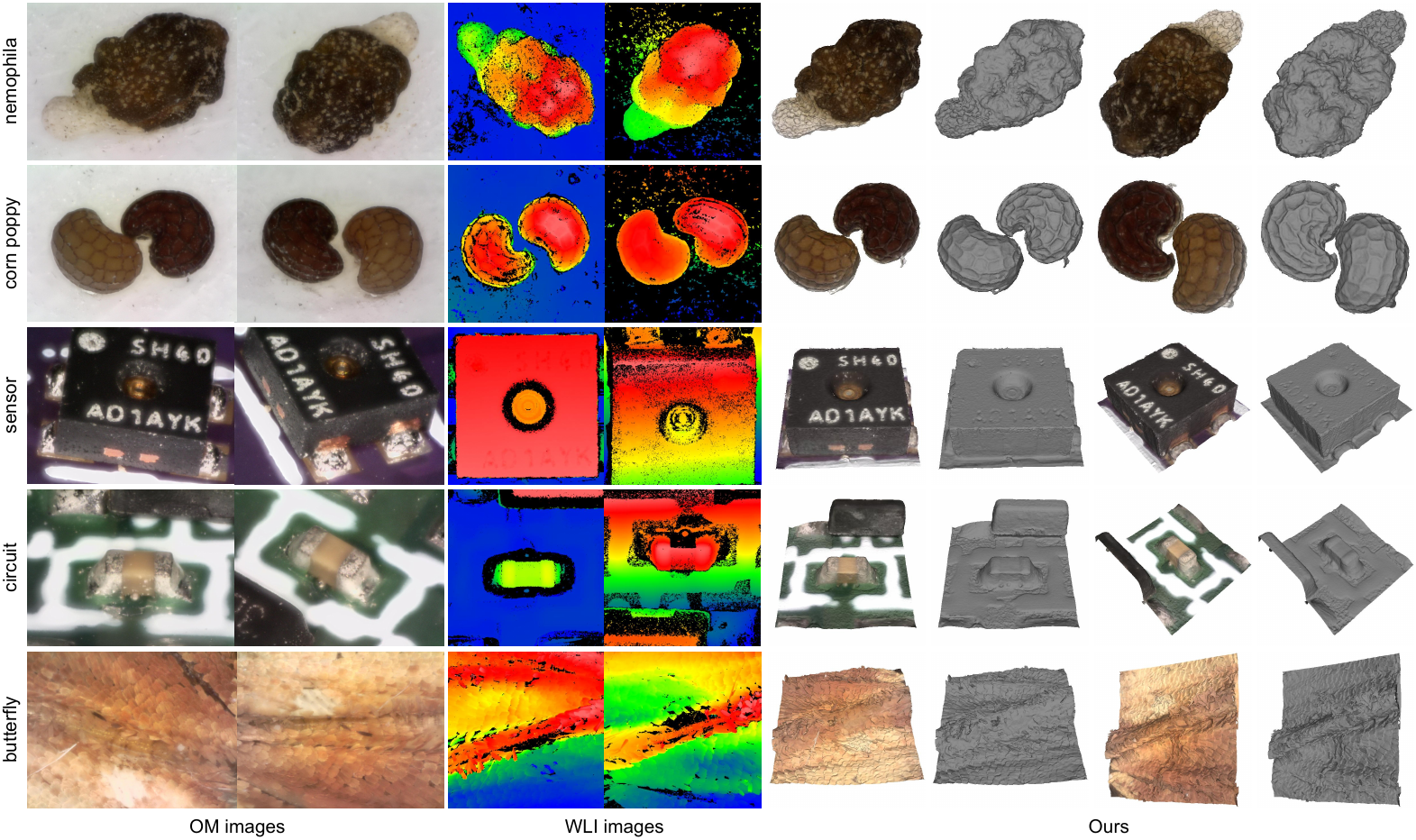}
    \caption{\textbf{More reconstruction results on the real-world WLI-OM dataset.} }
    \label{fig: more_real}
    \vspace{0em}
\end{figure*}

\begin{figure*}[h] 
    \centering
    \includegraphics[width=\linewidth]{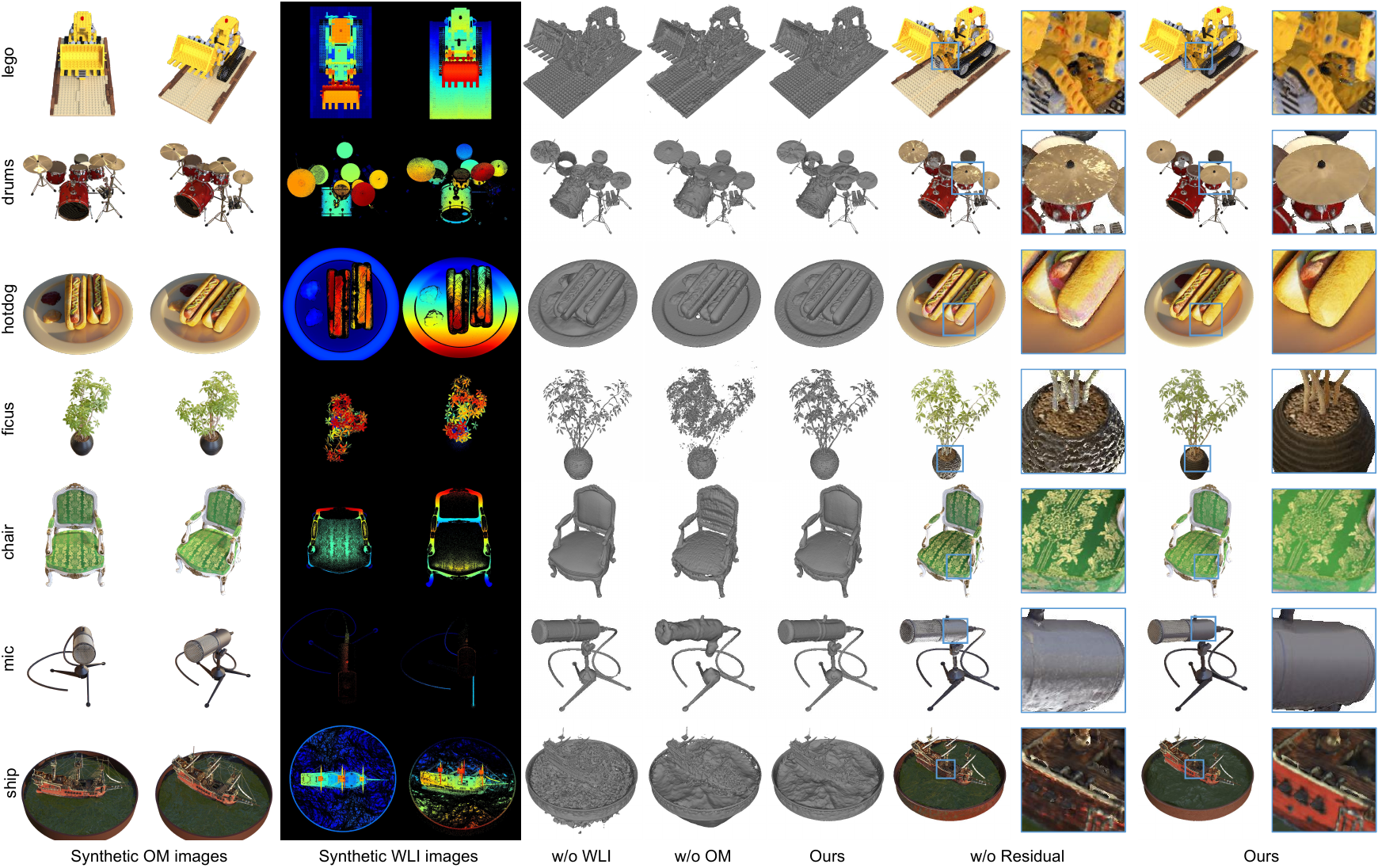}
    \caption{ \textbf{More reconstruction results on the synthetic multi-view WLI and OM dataset.} }
    \label{fig: more_sim}
    \vspace{-2em}
\end{figure*}

\end{document}